\documentclass{article}

\PassOptionsToPackage{numbers, compress}{natbib}
\usepackage[preprint]{neurips_2024}

\usepackage[utf8]{inputenc}
\usepackage[T1]{fontenc}
\usepackage{hyperref}
\usepackage{url}
\usepackage{booktabs}
\usepackage{array}
\usepackage{amsfonts}
\usepackage{amsmath}
\usepackage{amssymb}
\usepackage{nicefrac}
\usepackage{microtype}
\usepackage{xcolor}
\usepackage{graphicx}
\usepackage{float}
\usepackage{enumitem}
\usepackage{tikz}
\usetikzlibrary{arrows.meta,positioning,shapes.geometric,fit,backgrounds}

\let\svthefootnote\thefootnote
\newcommand\freefootnote[1]{%
	\let\thefootnote\relax%
	\footnotetext{#1}%
	\let\thefootnote\svthefootnote%
}

\title{DyssomniaNN: Multi-Scale Spiking Neural Modeling of Human Sleep Disorders and Perceived Restedness}

\author{%
  Daniel Szelogowski\\
  Department of Computer Science\\
  University of Wisconsin--Whitewater\\
  \texttt{szelogowdj19@uww.edu} \\
}

\begin{document}

\maketitle

\begin{abstract}
Clinical sleep medicine repeatedly documents profound divergences between objective sleep duration and subjective daytime impairment, notably exemplified by sleep state misperception in paradoxical insomnia and boundary instability in narcolepsy type 1. Here, we present \textbf{DyssomniaNN}: a multi-scale biophysical computational framework coupling Borb{\'e}ly's two-process regulation of sleep homeostasis and circadian drive with Saper's hypothalamic reciprocal flip-flop circuit and a recurrent corticothalamic leaky integrate-and-fire spiking neural network. We formalize five clinical cohorts across an identical 72-hour photoperiod: healthy control, circadian rhythm sleep-wake disorder (delayed-phase), chronic insomnia, paradoxical insomnia, and narcolepsy type 1. The synthetic local field potential converges closely with empirical human electroencephalography spectra from the PhysioNet Sleep-EDF database across all canonical frequency bands (spectral mean absolute error of $0.33$--$0.35$). Furthermore, we introduce the Well-Rested Index to quantify restorative efficacy. Simulations reveal that sleep duration fails to predict restorative recovery: paradoxical insomnia achieves normal total sleep time ($21.15\,\text{h}$) yet yields low restoration ($37.6\%$) from unsuppressed cortical beta-band intrusion, whereas hypocretin deficiency in narcolepsy produces severe nocturnal fragmentation ($781\,\text{min}$ of wake after sleep onset) and 35 sleep-onset rapid eye movement periods. DyssomniaNN provides an in silico platform for mechanistic sleep research and virtual neuropharmacology.

% Clinical sleep medicine repeatedly documents profound divergences between objective sleep duration and subjective daytime impairment, notably exemplified by sleep state misperception in paradoxical insomnia and boundary instability in narcolepsy type 1. Here, we present DyssomniaNN: a multi-scale biophysical computational framework coupling Borbély's two-process regulation of sleep homeostasis and circadian drive with Saper's hypothalamic reciprocal flip-flop circuit and a recurrent corticothalamic leaky integrate-and-fire spiking neural network. We formalize five clinical cohorts across an identical 72-hour photoperiod: healthy control, circadian rhythm sleep-wake disorder (delayed-phase), chronic insomnia, paradoxical insomnia, and narcolepsy type 1. The synthetic local field potential converges closely with empirical human electroencephalography spectra from the PhysioNet Sleep-EDF database across all canonical frequency bands (spectral mean absolute error of 0.33--0.35). Furthermore, we introduce the Well-Rested Index to quantify restorative efficacy. Simulations reveal that sleep duration fails to predict restorative recovery: paradoxical insomnia achieves normal total sleep time (21.15 h) yet yields low restoration (37.6%) from unsuppressed cortical beta-band intrusion, whereas hypocretin deficiency in narcolepsy produces severe nocturnal fragmentation (781 min of wake after sleep onset) and 35 sleep-onset rapid eye movement periods. DyssomniaNN provides an in silico platform for mechanistic sleep research and virtual neuropharmacology.

\freefootnote{Project source code available at \url{https://github.com/danielathome19/DyssomniaNN}.}
\end{abstract}

\section{Introduction}
\label{sec:introduction}

Sleep disorders affect an estimated $10\%$--$30\%$ of the global adult population, contributing substantially to cardiovascular disease, metabolic dysfunction, affective disorders, and cognitive decline \citep{berry2012aasm, riemann2010hyperarousal}. Clinical \textbf{polysomnography (PSG)} serves as the diagnostic benchmark for evaluating sleep pathology, categorizing sleep into discrete 30-second epochs comprising wakefulness, \textbf{non-rapid eye movement (NREM)} stages N1, N2, and N3 slow-wave sleep, and \textbf{rapid eye movement (REM)} sleep \citep{berry2012aasm}.

Despite the widespread clinical adoption of macroscale PSG metrics such as \textbf{total sleep time (TST)} and \textbf{sleep efficiency (SE)}, these indices frequently fail to correlate with patients' perceived daytime impairment or neurocognitive recovery \citep{edinger2003subtyping, krystal2002quantitative}. This divergence is evident in paradoxical insomnia, clinically designated as sleep state misperception. Patients suffering from paradoxical insomnia report profound subjective insomnia and daytime exhaustion, despite objective overnight PSG recording normal sleep duration, normal \textbf{sleep onset latency (SOL)}, and preserved macro-architecture \citep{edinger2003subtyping, krystal2002quantitative}. Conversely, patients diagnosed with narcolepsy type 1 frequently accumulate equal or greater TST than healthy individuals over a 24-hour cycle, yet suffer from severe daytime\linebreak \textbf{somnolence} (excessive daytime sleepiness and the inability to maintain consolidated wakefulness) and \textbf{cataplexy} (sudden, transient loss of voluntary muscle \textit{tone} --- i.e., steady neurotransmitter release --- triggered by strong emotions such as laughter, while consciousness remains preserved) \citep{dauvilliers2007narcolepsy, scammell2017narcolepsy}. These symptoms arise from the selective loss of hypothalamic neurons producing \textbf{hypocretin} and \textbf{orexin}, 
%(synonymous terms for the same neuropeptide family discovered independently and named hypocretins by de Lecea et al. and orexins by Sakurai et al.), 
which destabilizes transitions between wakefulness, NREM, and REM sleep \citep{peyron2000mutation, dauvilliers2007narcolepsy, scammell2017narcolepsy}.

Addressing these clinical discrepancies requires computational models capable of bridging microscopic synaptic and neuromodulatory mechanisms with macroscopic sleep architecture. Traditional computational models of sleep have largely existed in isolation: phenomenological differential equations modeling macroscopic circadian and homeostatic sleep propensity \citep{borbely1982two, daan1984timing, achermann1993two}, lumped-rate models of hypothalamic switching \citep{saper2001sleep, saper2010sleep, phillips2007sleep}, or biophysical \textbf{spiking neural networks (SNNs)} composed of \textbf{leaky integrate-and-fire (LIF)} neurons focused on local cortical oscillations during brief sleep spindles or slow waves \citep{destexhe1998mechanisms, gerstner2002spiking}. Prior modeling efforts have not united these biophysical scales into a continuous, multi-day simulation capable of reproducing distinct human dyssomnias within an identical environmental photoperiod (the structured daily cycle of ambient light and dark exposure, here configured as a 16-hour light/8-hour dark schedule).

In this work, we present \textbf{DyssomniaNN}, an end-to-end multi-scale biophysical computational framework implemented in PyTorch. Our contributions are summarized as follows:
\begin{enumerate}[itemsep=0.25em]
    \item We formulate a hierarchical biophysical model coupling Borb{\'e}ly's two-process regulation with Saper's reciprocal hypothalamic flip-flop switch and a corticothalamic LIF network across continuous multi-day photoperiods ($72.0\,\text{h}$).
    \item We parameterize five distinct clinical phenotypes: Healthy Normal Sleep, \textbf{Circadian Rhythm Sleep-Wake Disorder (CRSWD)} of the delayed-phase type, Chronic\linebreak Psychophysiological Insomnia (abbreviated in this paper as simply ``Chronic Insomnia''), Paradoxical Insomnia (Sleep State Misperception), and Narcolepsy Type 1.
    \item We validate emergent corticothalamic \textbf{electroencephalography (EEG)} spectra against empirical human PSG recordings from the PhysioNet Sleep-EDF database \citep{goldberger2000physiobank, kemp2000analysis}, achieving quantitative spectral convergence across Delta, Theta, Alpha, Sigma, and Beta bands.
    \item We formulate the \textbf{Well-Rested Index} ($\mathcal{R} \in [0, 100\%]$), demonstrating that perceived sleep restoration is dissociated from TST and is governed by the preservation of slow-wave sleep against cortical hyperarousal and circadian phase misalignment.
\end{enumerate}

\vspace{-0.25em}
\section{Related Work}
\label{sec:related_work}

\vspace{-0.5em}
Computational modeling of sleep regulation originates from the classic \textbf{Two-Process Model} formulated by Borb{\'e}ly and colleagues \citep{borbely1982two, daan1984timing}. The model posits that sleep timing and duration emerge from the nonlinear interaction between a homeostatic \textbf{sleep pressure process ($S$)}, which accumulates during wakefulness and discharges during sleep, and an oscillatory \textbf{circadian process ($C$)} driven by the \textbf{suprachiasmatic nucleus (SCN)} \citep{borbely1982two, daan1984timing}. While the two-process model provides a robust mathematical formulation of sleep debt and circadian gating, it operates strictly at the phenomenological level and lacks direct representation of the subcortical nuclei that execute state transitions.

Subsequent neurobiological investigations by Saper and colleagues identified the \textbf{ventrolateral preoptic nucleus (VLPO)} and \textbf{lateral hypothalamic area (LHA)} as the anatomical core of a mutual inhibitory flip-flop circuit governing sleep-wake transitions \citep{saper2001sleep, saper2010sleep, fuller2006neurobiology}. Inhibitory \textbf{GABAergic} and \textbf{galaninergic} projections (axonal pathways that transmit signals to distant brain regions) from the VLPO inhibit ascending \textbf{monoaminergic} and \textbf{peptidergic} arousal centers, including the locus coeruleus, dorsal raphe, tuberomammillary nucleus, and \textbf{orexinergic} neurons in the LHA \citep{saper2001sleep, saper2010sleep}. In contrast, excitatory \textbf{glutamatergic} and monoaminergic projections stimulate cortical activation and arousal pathways. Reciprocally, monoamines from the arousal systems inhibit the VLPO \citep{saper2001sleep, chamberlin2003adenosine}. Mathematical rate models of this reciprocal inhibition demonstrate that mutual inhibition generates \textbf{bistability} (a dynamical \textit{regime}/operating state possessing two distinct stable equilibria --- consolidated wake and consolidated sleep --- separated by an unstable boundary) and hysteresis, preventing intermediate, erratic state transitions in healthy physiology \citep{wilson1972excitatory, saper2010sleep, phillips2007sleep}. When orexinergic stabilization is lost, as observed in narcoleptic canine and human brains lacking prepro-orexin peptides \citep{peyron2000mutation, chemelli1999narcolepsy, dauvilliers2007narcolepsy, thannickal2000reduced}, the bistable switch undergoes rapid, involuntary transitions.

% \clearpage
At the cellular and network scale, computational neuroscience has explored the ionic and synaptic mechanisms underlying sleep rhythms using conductance-based Hodgkin-Huxley and LIF models \citep{destexhe1998mechanisms, gerstner2002spiking}. Thalamocortical loops produce low-frequency delta oscillations ($0.5$--$4.0\,\text{Hz}$) through hyperpolarization-activated cation currents (inward ionic currents carried by positively charged ions such as $\text{Na}^+$ and $\text{Ca}^{2+}$, specifically hyperpolarization-activated cyclic nucleotide-gated currents $I_h$ and low-threshold T-type calcium channels $I_T$) during deep N3 sleep \citep{destexhe1998mechanisms, stewart1993thalamocortical}. These currents interact to generate \textbf{sleep spindles} (transient $12.0$--$16.0\,\text{Hz}$ waxing-and-waning oscillatory bursts lasting $0.5$--$2.0\,\text{s}$ generated by reciprocal interactions between thalamic reticular interneurons and thalamocortical relay cells) during stage N2 sleep \citep{destexhe1998mechanisms, stewart1993thalamocortical}. In primary insomnia, quantitative EEG analyses demonstrate an elevated presence of high-frequency beta ($15.0$--$30.0\,\text{Hz}$) and gamma activity during NREM sleep \citep{perlis2001beta, krystal2002quantitative}, supporting the neurobiological hyperarousal model formulated by Riemann et al. \citep{riemann2010hyperarousal}. However, existing SNN models typically simulate isolated windows of several seconds or minutes, lacking coupling to continuous multi-day homeostatic and circadian drivers. DyssomniaNN bridges this division by driving recurrent corticothalamic LIF networks via continuous hypothalamic and homeostatic inputs over multi-day trajectories.

\section{Biophysical Framework}
\label{sec:biophysical_framework}

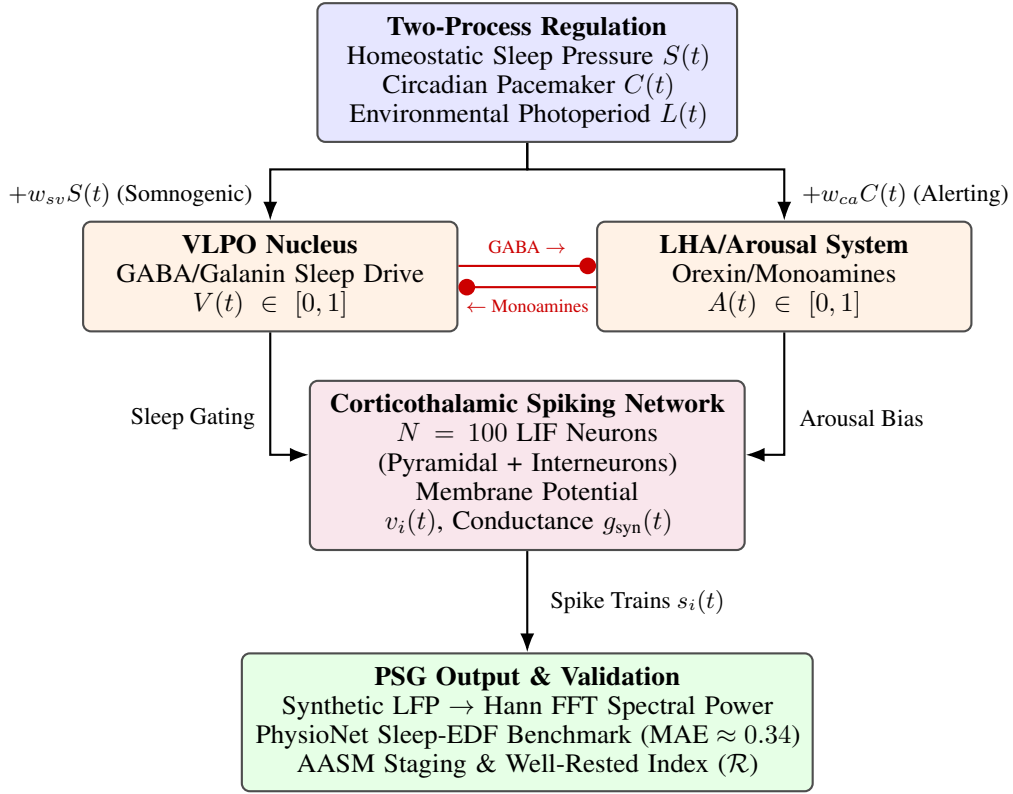
\begin{figure}[H]
\centering
\begin{tikzpicture}[
    scale=1.0, every node/.style={transform shape},
    box/.style={draw=black!70, rounded corners=3pt, thick, align=center, fill=white, inner sep=5pt},
    driver/.style={box, fill=blue!10, text width=5.2cm},
    switch/.style={box, fill=orange!10, text width=4.6cm},
    cortex/.style={box, fill=purple!10, text width=5.4cm},
    metric/.style={box, fill=green!10, text width=7.2cm},
    arrow/.style={-{Latex[length=2.5mm]}, thick},
    inhibit/.style={-{Circle[length=2.2mm]}, thick, red!80!black}
]

% Level 1: Two-Process Drivers
\node[driver] (sc) at (0, 4.6) {
    \textbf{Two-Process Regulation}\\
    Homeostatic Sleep Pressure $S(t)$\\
    Circadian Pacemaker $C(t)$\\
    Environmental Photoperiod $L(t)$
};

% Level 2: Hypothalamic Switch
\node[switch] (vlpo) at (-3.4, 1.9) {
    \textbf{VLPO Nucleus}\\
    GABA/Galanin Sleep Drive\\
    $V(t)\!\in\![0, 1]$
};

\node[switch] (lha) at (3.4, 1.9) {
    \textbf{LHA/Arousal System}\\
    Orexin/Monoamines\\
    $A(t)\!\in\![0, 1]$
};

% Level 3: Corticothalamic Network
\node[cortex] (snn) at (0, -0.6) {
    \textbf{Corticothalamic Spiking Network}\\
    $N\!=\!100$ LIF Neurons (Pyramidal + Interneurons)\\
    Membrane Potential $v_i(t)$, Conductance $g_{\text{syn}}(t)$
};

% Level 4: Validation & Metrics (widened box to prevent awkward line breaks)
\node[metric] (output) at (0, -4.0) {
    \textbf{PSG Output \& Validation}\\
    Synthetic LFP $\to$ Hann FFT Spectral Power\\
    PhysioNet Sleep-EDF Benchmark ($\text{MAE} \approx 0.34$)\\
    AASM Staging \& Well-Rested Index ($\mathcal{R}$)
};

% Connections: Drivers to Hypothalamic Switch (annotated with clear biophysical meaning)
\draw[arrow] (sc.south) -- ++(0,-0.35) -| (vlpo.north) node[pos=0.75, left=3pt, font=\footnotesize] {$+w_{sv} S(t)$ (Somnogenic)};
\draw[arrow] (sc.south) -- ++(0,-0.35) -| (lha.north) node[pos=0.75, right=3pt, font=\footnotesize] {$+w_{ca} C(t)$ (Alerting)};

% Reciprocal Inhibition between VLPO and LHA
\draw[inhibit] ([yshift=4pt]vlpo.east) -- ([yshift=4pt]lha.west) 
    node[midway, above=1pt, font=\scriptsize, text=red!80!black] {GABA $\to$};
\draw[inhibit] ([yshift=-4pt]lha.west) -- ([yshift=-4pt]vlpo.east) 
    node[midway, below=1pt, font=\scriptsize, text=red!80!black] {$\leftarrow$ Monoamines};

% Hypothalamus to Corticothalamic Network
\draw[arrow] (vlpo.south) |- ([yshift=4pt]snn.west) node[pos=0.35, left=2pt, font=\footnotesize] {Sleep Gating};
\draw[arrow] (lha.south) |- ([yshift=4pt]snn.east) node[pos=0.35, right=2pt, font=\footnotesize] {Arousal Bias};

% SNN to Polysomnographic Output
\draw[arrow] (snn.south) -- (output.north) node[midway, right=5pt, font=\footnotesize] {Spike Trains $s_i(t)$};

\end{tikzpicture}
\caption{Architectural topology of the DyssomniaNN framework. Borb{\'e}ly's two-process regulators ($S(t)$ and $C(t)$) drive the reciprocal hypothalamic flip-flop switch (VLPO $\leftrightarrow$ LHA), where $+w_{sv} S(t)$ represents homeostatic somnogenic excitation of the VLPO and $+w_{ca} C(t)$ represents circadian alerting excitation of the LHA. The resulting subcortical arousal level gates the corticothalamic LIF network, whose synthetic LFP is evaluated against empirical PSG spectra from the PhysioNet Sleep-EDF database.}
\label{fig:architecture}
\end{figure}

The architecture of DyssomniaNN comprises three interconnected hierarchical layers depicted in \underline{Figure~\ref{fig:architecture}}: homeostatic-circadian regulation, hypothalamic bistable switching, and a recurrent corticothalamic SNN.

\vspace{2em}

% \clearpage
\subsection{Two-Process Homeostatic and Circadian Model}
\label{sec:process_sc}

Homeostatic sleep pressure $S(t) \in [0, 1]$ represents the neurochemical accumulation and clearance of endogenous somnogens (naturally occurring biochemical compounds synthesized within the central nervous system that promote sleepiness --- primarily extracellular adenosine, inflammatory cytokines, and prostaglandin D2). Neurochemical assays (in vivo microdialysis and electrochemical biosensor measurements of interstitial brain fluid) demonstrate that extracellular adenosine accumulates in the basal forebrain and cerebral cortex during wakefulness as a consequence of sustained metabolic consumption of \textbf{adenosine triphosphate (ATP)} by active neurons and astrocytes (star-shaped non-neuronal glial cells that regulate synaptic transmission, clear neurotransmitters, and release adenosine via gliotransmission) \citep{porkka1997adenosine, benington1995restoration}. During sleep, astrocytic reuptake and enzymatic breakdown by adenosine deaminase rapidly reduce extracellular adenosine concentrations \citep{porkka1997adenosine}. Following the classical Two-Process formulation \citep{borbely1982two, daan1984timing}, the continuous dynamics of $S(t)$ are modeled by coupled first-order differential equations:
\begin{equation}
\frac{dS(t)}{dt} = 
\begin{cases}
\frac{1 - S(t)}{\tau_r}, & \text{if awake } (M(t) = 0), \\
-\frac{S(t)}{\tau_d}, & \text{if asleep } (M(t) > 0),
\end{cases}
\label{eq:process_s}
\end{equation}
where $M(t) \in \{0, 1, 2, 3, 4\}$ denotes the discrete PSG stage at time $t$ (corresponding to Wake, N1, N2, N3, and REM, respectively). During wakefulness ($M(t) = 0$), $S(t)$ asymptotically approaches an upper saturation limit ($S_{\text{max}} = 1.0$) with a rise time constant $\tau_r = 18.2\,\text{h}$. During sleep ($M(t) > 0$), $S(t)$ decays exponentially toward zero with a clearance time constant $\tau_d = 4.2\,\text{h}$. These empirical time constants were derived by Daan, Beersma, and Borb{\'e}ly from human \textbf{slow-wave activity (SWA)} saturation curves across extended 40-hour sleep deprivation and subsequent recovery protocols \citep{daan1984timing, achermann1993two}.

The circadian alerting drive $C(t)$ is generated by cell-autonomous transcriptional-translational feedback loops within the SCN, exhibiting an intrinsic circadian period near 24.2 hours in humans \citep{fuller2006neurobiology, sack2007circadian}. The alerting drive is modeled as a periodic oscillator coupled to external environmental lux $L(t)$ (equal to one lumen per square meter, measuring perceived ambient luminous intensity):
\begin{equation}
C(t) = A \sin\left(\frac{2\pi (t - t_0 - \Delta\phi)}{24}\right) + \kappa \frac{L(t)}{L_{\text{max}} + L(t)},
\label{eq:process_c}
\end{equation}
where $A = 0.38$ represents the normalized circadian oscillation amplitude calibrated by Daan et al. \citep{daan1984timing} and Achermann and Borb{\'e}ly \citep{achermann1993two} to counterbalance waking homeostatic accumulation across an ordinary 16-hour waking day.

The parameter $t_0 = 6.0\,\text{h}$ sets the \textbf{diurnal nadir} (the lowest point of the 24-hour cycle) of the circadian wake-maintenance signal at approximately 05:00--06:00, corresponding to the physiological minimum of human core body temperature and alertness \citep{daan1984timing, zeitzer2000sensitivity}. The parameter $\Delta\phi$ denotes the endogenous circadian phase offset in hours, allowing formal modeling of chronotypes and phase-shift disorders \citep{sack2007circadian}. External photic entrainment is mediated by \textbf{intrinsically photosensitive retinal ganglion cells (ipRGCs)} expressing melanopsin, which project monosynaptically to the SCN via the retinohypothalamic tract \citep{berson2002phototransduction}. Photic saturation is formalized via a saturating Hill-type function with a half-saturation constant $L_{\text{max}} = 10{,}000\,\text{lux}$ and coupling gain $\kappa = 0.12$, matching empirical human phase-resetting curves and melatonin suppression dynamics \citep{zeitzer2000sensitivity, phillips2007sleep}.

\subsection{Reciprocal Hypothalamic Flip-Flop Dynamics}
\label{sec:flip_flop}

State transitions between sleep and wakefulness are executed by the reciprocal inhibitory flip-flop circuit identified by Saper and colleagues \citep{saper2001sleep, saper2010sleep}. The circuit comprises two mutually antagonistic subcortical populations: the sleep-active VLPO, characterized by normalized mean firing rate $V(t) \in [0, 1]$, and the wake-active ascending monoaminergic and lateral hypothalamic arousal systems, characterized by normalized mean firing rate $A(t) \in [0, 1]$ (both dimensionless population rates scaled from 0, denoting complete \textit{quiescence} (inactivity), to 1, denoting maximal saturating discharge of approximately $10\text{--}20\,\text{Hz}$) \citep{saper2001sleep, fuller2006neurobiology}. The population dynamics are \linebreak formalized using the classical \textbf{Wilson-Cowan mean-field formulation} \citep{wilson1972excitatory, phillips2007sleep}:
\begin{align}
\tau_v \frac{dV(t)}{dt} &= -V(t) + \sigma_\beta\left(w_{sv} S(t) - w_{cv} C(t) - w_{av} A(t) + \theta_v\right), \label{eq:vlpo} \\
\tau_a \frac{dA(t)}{dt} &= -A(t) + \sigma_\beta\left(w_{ca} C(t) + \omega_{\text{orexin}} \cdot \Omega_{\text{tone}} - w_{va} V(t) + I_{\text{arousal}} + \theta_a\right), \label{eq:lha}
\end{align}
where $\tau_v = 1.8\,\text{min}$ and $\tau_a = 1.2\,\text{min}$ represent the macroscopic integration timescales of subcortical monoaminergic and peptidergic volume transmission, which operate over minutes rather than milliseconds \citep{phillips2007sleep}. The function $\sigma_\beta(u) = (1 + e^{-\beta u})^{-1}$ is a sigmoidal activation function whose steepness $\beta = 5.0$ reflects the cooperative threshold firing response of hypothalamic neuronal ensembles \citep{wilson1972excitatory, chamberlin2003adenosine}.

% \clearpage
The physiological inputs to the system are grounded in known neuroanatomy:  % TODO: revise, this is a bold claim and pointless
\begin{itemize}
    \item \textbf{Homeostatic Somnogenic Drive} ($+w_{sv} S(t)$, $w_{sv} = 1.6$): Extracellular adenosine excites VLPO neurons directly via postsynaptic $A_{2A}$ receptors (excitatory G-protein coupled receptors) and disinhibits the VLPO (relieving target neurons from upstream inhibition) by suppressing presynaptic GABAergic inputs via inhibitory $A_1$ receptors \citep{porkka1997adenosine, chamberlin2003adenosine}.
    \item \textbf{Circadian Alerting Drive} ($-w_{cv} C(t)$ and $+w_{ca} C(t)$, $w_{cv} = 0.9, w_{ca} = 1.1$): Efferent projections from the SCN relay through the \textbf{dorsomedial hypothalamic nucleus (DMH)} to stimulate wake-promoting LHA orexinergic and monoaminergic nuclei during subjective day, while simultaneously providing indirect GABAergic inhibition to the VLPO \citep{saper2001sleep, fuller2006neurobiology}.
    \item \textbf{Orexinergic Wake Stabilization} ($+\omega_{\text{orexin}} \cdot \Omega_{\text{tone}}$, $\omega_{\text{orexin}} = 1.5$): Peptidergic projections from hypocretin/orexin neurons in the LHA project excitatory inputs to monoaminergic nuclei (locus coeruleus, dorsal raphe, and tuberomammillary nucleus), stabilizing the wake-active attractor \citep{peyron2000mutation, saper2001sleep, scammell2017narcolepsy}. $\Omega_{\text{tone}} \in [0, 1]$ represents the intactness of endogenous hypocretin peptide availability ($\Omega_{\text{tone}} = 1.0$ in intact physiology; $\Omega_{\text{tone}} \approx 0$ under complete neurodegeneration).
    \item \textbf{Hyperarousal Bias} ($I_{\text{arousal}} \in \mathbb{R}$): Represents ascending depolarizing drive from the limbic system, prefrontal cortex, and sympathetic axis induced by psychological stress or conditioned cognitive arousal \citep{riemann2010hyperarousal}.
    \item \textbf{Basal Rheobase Thresholds} ($\theta_v = -0.5, \theta_a = -0.3$): Ensure that in the absence of net positive synaptic input, baseline neuronal membrane potentials remain subthreshold and quiescent \citep{phillips2007sleep}.
\end{itemize}

\paragraph{Mathematical Derivation of Bistability and Hysteresis.}
The mutual inhibitory weights $w_{av} = 2.1$ and $w_{va} = 2.0$ are derived from bifurcation analysis of the two-variable Wilson-Cowan system \citep{wilson1972excitatory, phillips2007sleep}. The nullclines of \underline{Equations~\eqref{eq:vlpo} and \eqref{eq:lha}} are obtained by setting $dV/dt = 0$ and $dA/dt = 0$:
\begin{align}
V &= \sigma_\beta\left(w_{sv} S(t) - w_{cv} C(t) - w_{av} A + \theta_v\right), \label{eq:nullcline_v} \\
A &= \sigma_\beta\left(w_{ca} C(t) + \omega_{\text{orexin}} \Omega_{\text{tone}} - w_{va} V + I_{\text{arousal}} + \theta_a\right). \label{eq:nullcline_a}
\end{align}
In a mutual inhibitory dynamical system with sigmoidal activation functions $\sigma_\beta(u)$, bistability arises through a saddle-node or pitchfork bifurcation when mutual inhibitory coupling becomes sufficiently strong to split a single central uncommitted state into two stable steady states \citep{wilson1972excitatory, phillips2007sleep}. Because $\max(\sigma_\beta'(u)) = \beta/4$, the necessary criterion for bistability requires that the product of the mutual inhibitory gains exceeds the inverse product of the maximum slopes of the activation functions:
\begin{equation}
w_{av} \cdot w_{va} > \frac{4}{\beta}.
\label{eq:bistability_condition}
\end{equation}
For our population steepness $\beta = 5.0$, the critical bifurcation threshold is $4/\beta = 0.8$. With $w_{av} = 2.1$ and $w_{va} = 2.0$, the interaction gain is $w_{av} \cdot w_{va} = 4.2 \gg 0.8$. This configuration places the hypothalamic switch firmly within the bistable regime, establishing two distinct stable attractors: a consolidated wake state ($A \approx 1, V \approx 0$) and a consolidated sleep state ($V \approx 1, A \approx 0$), separated by an unstable threshold manifold. This non-equilibrium bistability generates wide \textbf{hysteresis} (the dependence of the state of the system on its prior history, meaning greater excitatory drive is needed to switch states than to maintain an existing state), ensuring that behavioral transitions are decisive and resistant to baseline fluctuations \citep{saper2001sleep, saper2010sleep}. When endogenous hypocretin peptide concentration $\Omega_{\text{tone}}$ is depleted, the wake attractor's basin of attraction shrinks, shifting the system toward the bifurcation boundary where small perturbations trigger involuntary state transitions \citep{chemelli1999narcolepsy, scammell2017narcolepsy}.

\subsection{Corticothalamic Spiking Neural Network}
\label{sec:snn}

Cortical electrodynamics are simulated via a recurrent population of $N = 100$ point neurons adhering to Dale's principle: $80\%$ excitatory pyramidal neurons ($N_E = 80$) and $20\%$ inhibitory GABAergic fast-spiking interneurons ($N_I = 20$), preserving canonical neocortical cytoarchitecture \citep{mccormick1985comparative, song2005highly}. The membrane potential $v_i(t)$ of neuron $i$ is modeled using LIF dynamics derived from a parallel resistor-capacitor (RC) equivalent circuit rather than full multi-conductance Hodgkin-Huxley equations, balancing biophysical realism with computational tractability over multi-day continuous simulations \citep{gerstner2002spiking}:
\begin{equation}
C_m \frac{dv_i(t)}{dt} = -g_L (v_i(t) - v_{\text{rest}}) + I_i^{\text{syn}}(t) + I_i^{\text{drive}}(t) + \eta_i(t).
\label{eq:rc_circuit}
\end{equation}
Dividing by leak conductance $g_L$ yields the standard LIF formulation:
\begin{equation}
\tau_m \frac{dv_i(t)}{dt} = -(v_i(t) - v_{\text{rest}}) + R_m \left( I_i^{\text{syn}}(t) + I_i^{\text{drive}}(t) + \eta_i(t) \right),
\label{eq:lif}
\end{equation}
where $\tau_m = C_m/g_L = R_m C_m = 20.0\,\text{ms}$ is the membrane time constant of cortical pyramidal cells in vitro (i.e., in a laboratory setting) \citep{mccormick1985comparative, destexhe2003fluctuating}. The membrane resistance is $R_m = 1.0\,\text{G}\Omega$ (reflecting the input resistance of cortical pyramidal cells in low-conductance states), resting potential $v_{\text{rest}} = -65.0\,\text{mV}$, action potential threshold $v_{\text{th}} = -50.0\,\text{mV}$, reset potential $v_{\text{reset}} = -70.0\,\text{mV}$, and absolute refractory period $\tau_{\text{ref}} = 2.0\,\text{ms}$, in agreement with intracellular patch-clamp measurements of regular-spiking neocortical pyramidal neurons \citep{mccormick1985comparative, destexhe2003fluctuating}. The term $\eta_i(t) \sim \mathcal{N}(0, \sigma^2)$ is Gaussian white noise with $\sigma = 1.5\,\text{pA}$, modeling continuous stochastic synaptic bombardment and thermal channel fluctuations in vivo (i.e., living brain tissue) \citep{destexhe2003fluctuating}. When $v_i(t) \ge v_{\text{th}}$, the neuron emits a spike $s_i(t) = 1$, and its membrane potential is clamped to $v_{\text{reset}}$ for $\tau_{\text{ref}}$.

Recurrent synaptic weights $W_{ij}$ for excitatory connections are sampled from a log-normal distribution:
\begin{equation}
W_{ij} \sim \text{LogNormal}(\mu_{\ln} = -1.2, \sigma_{\ln} = 0.6), \quad \text{for } j \in E,
\label{eq:lognormal}
\end{equation}
where $E = \{1, \dots, N_E\}$ denotes the set of excitatory pyramidal neurons, reflecting the heavy-tailed, non-random synaptic connectivity documented across mammalian cortical microcircuits \citep{song2005highly}. Synaptic currents follow a difference-of-exponentials formulation to capture finite rise and decay times:
\begin{equation}
I_i^{\text{syn}}(t) = \sum_{j=1}^N W_{ij} \int_0^\infty \left( \frac{e^{-s/\tau_{\text{decay}}} - e^{-s/\tau_{\text{rise}}}}{\tau_{\text{decay}} - \tau_{\text{rise}}} \right) s_j(t - s) \, ds,
\label{eq:synapse}
\end{equation}
where excitatory \textbf{$\alpha$-amino-3-hydroxy-5-methyl-4-isoxazolepropionic acid (AMPA)} receptor-mediated transmission (fast glutamate-gated ion channels) is parameterized by $\tau_{\text{rise}} = 0.5\,\text{ms}$ and $\tau_{\text{decay}} = 2.0\,\text{ms}$ \citep{destexhe1998mechanisms, destexhe2003fluctuating}. Inhibitory $\text{GABA}_A$ receptor-mediated transmission is parameterized by $\tau_{\text{rise}} = 1.0\,\text{ms}$ and $\tau_{\text{decay}} = 6.0\,\text{ms}$ \citep{destexhe1998mechanisms}, scaled by the cohort-specific cortical inhibition factor $\gamma_{\text{inh}}$ ($1.0$ in controls, attenuated in paradoxical insomnia) \citep{plante2012reduced}.

\paragraph{Neuromodulatory and Thalamic Gating Drive.}
Subcortical ascending projections gate the operating regime of the corticothalamic network between waking desynchrony and slow-wave synchrony \citep{stewart1993thalamocortical, destexhe1998mechanisms}. This input is formalized as:
\begin{equation}
I_i^{\text{drive}}(t) = I_0 \cdot A(t) + I_{\text{slow}} \cdot V(t) \cdot \cos(2\pi f_{\text{slow}} t).
\label{eq:drive}
\end{equation}
During wakefulness ($A(t) \approx 1, V(t) \approx 0$), ascending monoaminergic and cholinergic projections depolarize cortical pyramidal and thalamic relay neurons via inhibition of leak potassium conductances ($I_{\text{K,leak}}$) \citep{mccormick1985comparative, stewart1993thalamocortical}. The tonic depolarizing current $I_0 \cdot A(t)$ ($I_0 = 12.0\,\text{pA}$) maintains the network in an asynchronous, irregular, high-conductance state characteristic of wakefulness \citep{destexhe2003fluctuating}. During NREM sleep ($V(t) \approx 1, A(t) \approx 0$), withdrawal of subcortical monoaminergic depolarization hyperpolarizes thalamocortical relay cells. Hyperpolarization deinactivates low-threshold T-type $\text{Ca}^{2+}$ currents ($I_T$, transient voltage-gated calcium currents that activate at hyperpolarized potentials to produce burst firing) and hyperpolarization-activated cyclic nucleotide-gated cation currents ($I_h$), producing rhythmic burst-firing at slow oscillation frequencies ($f_{\text{slow}} = 1.0\,\text{Hz}$) \citep{stewart1993thalamocortical, destexhe1998mechanisms}. This slow thalamic rhythmic pacemaker current $I_{\text{slow}} \cdot V(t) \cdot \cos(2\pi f_{\text{slow}} t)$ ($I_{\text{slow}} = 15.0\,\text{pA}$) entrains cortical neurons into synchronized slow-wave Up/Down state alternations \citep{stewart1993thalamocortical, buzsaki2012origin}.

\subsection{Synthetic Local Field Potential and Spectral Integration}
\label{sec:lfp}

The aggregate extracellular field potential recorded during clinical PSG reflects transmembrane currents flowing in cortical tissue \citep{buzsaki2012origin}. Classical volume conductor theory establishes the physical principles governing how transmembrane currents propagate through conductive brain tissue to generate extracellular electrical potentials \citep{nunez2006electric, buzsaki2012origin}. Fast action potentials produce an inward sodium current flanked symmetrically by outward potassium currents, creating an opposing quadrupolar current configuration. The electric fields of these quadrupoles decay rapidly with distance as $1/r^3$, resulting in extensive spatial cancellation and negligible contributions to macroscopic recordings. In contrast, synchronized postsynaptic currents flowing into the aligned apical dendrites of layer V pyramidal neurons (large excitatory projection neurons whose elongated dendrites span multiple cortical layers perpendicular to the brain surface) create an open sink-source current configuration that acts as a current dipole. The electric fields of current dipoles decay more slowly as $1/r^2$ and sum coherently across large cortical assemblies, dominating both the \textbf{local field potential (LFP)} recorded within tissue and scalp EEG signals \citep{nunez2006electric, buzsaki2012origin, mazzoni2008computational}.

Accordingly, the synthetic LFP is modeled by convolving the network's aggregate spiking events with an alpha-function postsynaptic dendritic response kernel $h(t)$:
\begin{equation}
\text{LFP}(t) = \frac{1}{N} \sum_{i=1}^N \int_0^\infty h(\tau) s_i(t - \tau) \, d\tau, \quad h(t) = \frac{t}{\tau_{\text{dend}}} e^{-t/\tau_{\text{dend}}},
\label{eq:lfp}
\end{equation}
where $\tau_{\text{dend}} = 4.0\,\text{ms}$ encapsulates dendritic cable low-pass filtering and the extracellular resistive-capacitive impedance of cortical tissue \citep{buzsaki2012origin, mazzoni2008computational}.

% \clearpage
The continuous synthetic LFP is sampled at $f_s = 256\,\text{Hz}$ across non-overlapping 1.0-second epochs ($K = 256$ samples per window), adhering to clinical digital EEG recording standards \citep{kemp2000analysis, berry2012aasm}. To mitigate spectral leakage (spurious frequency artifacts caused by discontinuities at finite epoch boundaries in discrete Fourier analysis), each discrete epoch $\text{LFP}[k]$ is multiplied by a symmetric Hann window taper $w[k]$ (a smooth raised-cosine taper that approaches zero at both window edges):
\begin{equation}
w[k] = 0.5 - 0.5 \cos\left(\frac{2\pi k}{K - 1}\right), \quad 0 \le k \le K - 1.
\label{eq:hann}
\end{equation}
The tapered signal is transformed into the frequency domain via the discrete \textbf{Fast Fourier Transform (FFT)} --- $\mathcal{F}\{\text{LFP} \cdot w\}(f)$. Spectral band power is integrated across the five canonical clinical frequency bands defined by the \textbf{American Academy of Sleep Medicine (AASM)} \citep{berry2012aasm}:
\begin{equation}
P_{\text{band}} = \int_{f_{\text{low}}}^{f_{\text{high}}} |\mathcal{F}\{\text{LFP} \cdot w\}(f)|^2 \, df,
\label{eq:band_power}
\end{equation}
spanning the following clinical frequency bands:
\begin{itemize}[itemsep=0.5pt, parsep=0pt, topsep=1.0pt, partopsep=0pt]
    \item \textbf{Delta} ($0.5$--$4.0\,\text{Hz}$): High-amplitude slow-wave activity dominant during deep stage N3 slow-wave sleep.
    \item \textbf{Theta} ($4.0$--$8.0\,\text{Hz}$): Synchronized rhythm characteristic of drowsiness, light stage N1 sleep, and REM sleep.
    \item \textbf{Alpha} ($8.0$--$13.0\,\text{Hz}$): Prominent occipital rhythm associated with relaxed wakefulness with eyes closed.
    \item \textbf{Sigma} ($12.0$--$16.0\,\text{Hz}$): Thalamocortical sleep spindle activity characteristic of stage N2 sleep.
    \item \textbf{Beta} ($15.0$--$30.0\,\text{Hz}$): Desynchronized, low-amplitude activity reflecting alertness, cognitive engagement, and cortical hyperarousal.
\end{itemize}
Relative power is computed by normalizing against total spectral power: $\bar{P}_{\text{band}} = P_{\text{band}}/\sum_{b \in \mathcal{B}} P_b$.

\section{Cohort Parameterization and Disorder Phenotypes}
\label{sec:cohorts}

\begin{table}[H]
\caption{Biophysical parameterization across the five simulated clinical cohorts. Values departing from normal control physiology are bolded and grounded in empirical clinical findings \citep{sack2007circadian, uchiyama2000poor, riemann2010hyperarousal, krystal2002quantitative, perlis2001beta, scammell2017narcolepsy, dauvilliers2007narcolepsy}.}
\label{tab:cohort_params}
\centering
\setlength{\tabcolsep}{4.5pt}
\small
\begin{tabular}{lccccc}
\toprule
\textbf{Biophysical Parameter} & \textbf{Normal} & \textbf{CRSWD} & \textbf{Chronic Insomnia} & \textbf{Paradoxical} & \textbf{Narcolepsy} \\
\midrule
Phase Offset $\Delta\phi$ (h) & $0.0$ & $\mathbf{+4.0}$ & $0.0$ & $0.0$ & $0.0$ \\
Circadian Amplitude $A$ & $0.38$ & $\mathbf{0.20}$ & $0.38$ & $0.38$ & $0.38$ \\
Orexin Tone $\Omega_{\text{tone}}$ & $1.0$ & $1.0$ & $1.0$ & $1.0$ & $\mathbf{0.05}$ \\
Arousal Bias $I_{\text{arousal}}$ (pA) & $0.0$ & $0.0$ & $\mathbf{+0.60}$ & $0.0$ & $-0.10$ \\
VLPO Inhibitory Weight $w_{av}$ & $2.1$ & $2.1$ & $\mathbf{0.95}$ & $2.1$ & $2.1$ \\
Cortical Inh. Scale $\gamma_{\text{inh}}$ & $1.0$ & $1.0$ & $1.0$ & $\mathbf{0.25}$ & $1.0$ \\
REM Entry Latency $\theta_{\text{rem}}$ (min) & $70.0$ & $70.0$ & $70.0$ & $70.0$ & $\mathbf{5.0}$ \\
\bottomrule
\end{tabular}
\end{table}

\vspace{1em}
To simulate clinically recognized dyssomnias, empirical pathophysiological deficits are mapped onto the framework's biophysical parameters, as summarized in \underline{Table~\ref{tab:cohort_params}}.

\paragraph{Cohort 0: Healthy Normal Control.}
Represents unperturbed sleep homeostasis (the biological process regulating the accumulation of somnogenic sleep drive during wakefulness and its exponential discharge during sleep) and circadian entrainment (the synchronization of endogenous circadian oscillations to the 24-hour environmental photoperiod) \citep{borbely1982two, fuller2006neurobiology}. Circadian phase is synchronized with the external photoperiod ($\Delta\phi = 0.0\,\text{h}$), and circadian amplitude reflects standard homeostatic balance ($A = 0.38$) \citep{daan1984timing, achermann1993two}. Hypothalamic flip-flop parameters satisfy the bistability criterion ($w_{av} = 2.1, w_{va} = 2.0$, yielding $w_{av} w_{va} = 4.2 > 0.8$), orexin tone is intact ($\Omega_{\text{tone}} = 1.0$, corresponding to normal \textbf{cerebrospinal fluid (CSF)} orexin-A concentrations $\ge 200\,\text{pg/mL}$ \citep{dauvilliers2007narcolepsy}), arousal bias is neutral ($I_{\text{arousal}} = 0.0\,\text{pA}$), cortical inhibition is intact ($\gamma_{\text{inh}} = 1.0$), and REM entry latency adheres to the physiological NREM cycle duration ($\theta_{\text{rem}} = 70.0\,\text{min}$) \citep{feinberg1979systematic, carskadon2011monitoring}.

\paragraph{Cohort 1: Delayed Sleep-Wake Phase Disorder (CRSWD).}
Captures intrinsic circadian dysregulation characterized by a delayed endogenous phase angle. In clinical cohorts, patients exhibit a 3 to 6-hour delay in the timing of \textbf{dim light melatonin onset (DLMO)} and core body temperature nadir \citep{sack2007circadian, uchiyama2000poor}. This is formalized by setting $\Delta\phi = +4.0\,\text{h}$. Furthermore, clinical studies document an attenuated circadian pacemaking amplitude in delayed phase syndrome \citep{sack2007circadian, uchiyama2000poor}, parameterized here as an approximate $50\%$ reduction in circadian amplitude ($A = 0.20$), which impairs morning wakefulness and delays evening somnolence (the physiological drive toward sleep).

\paragraph{Cohort 2: Chronic [Psychophysiological] Insomnia.}
Models conditioned cognitive and autonomic hyperarousal \citep{riemann2010hyperarousal}. Chronic insomnia patients exhibit elevated sympathetic nervous activity, increased nocturnal cortisol secretion, and higher whole-body metabolic rates \citep{riemann2010hyperarousal}. In DyssomniaNN, this allostatic load (the cumulative physiological wear and tear resulting from sustained neuroendocrine and autonomic hyperactivation) is parameterized by a chronic depolarizing arousal bias ($I_{\text{arousal}} = +0.60\,\text{pA}$). Concurrently, subcortical sleep-inducing strength is compromised by reducing VLPO inhibitory efficacy to $w_{av} = 0.95$ \citep{saper2010sleep, riemann2010hyperarousal}. Because $w_{av} \cdot w_{va} = 0.95 \times 2.0 = 1.9$, the flip-flop switch remains above the critical bifurcation point ($0.8$) but operates close to the boundary, making the sleep attractor (the stable equilibrium state corresponding to consolidated sleep in phase space) shallow and vulnerable to spontaneous awakenings.

\paragraph{Cohort 3: Paradoxical Insomnia (Sleep State Misperception).}
Characterized by a dissociation between subcortical sleep staging and cortical spectral activity \citep{edinger2003subtyping, krystal2002quantitative}. Subcortical hypothalamic parameters remain identical to the healthy control cohort ($\Delta\phi = 0.0\,\text{h}, \Omega_{\text{tone}} = 1.0, w_{av} = 2.1$). However, \textbf{proton magnetic resonance spectroscopy ($^1\text{H}$-MRS)} studies in insomnia patients document substantial ($20\%$--$30\%$) reductions in cortical GABA concentrations \citep{plante2012reduced}. In our model, this is formalized by scaling cortical interneuron synaptic gain down to $\gamma_{\text{inh}} = 0.25$. This selective cortical disinhibition prevents the suppression of fast recurrent synaptic currents during slow-wave sleep, generating pathological high-frequency beta-band intrusion ($15.0$--$30.0\,\text{Hz}$) while preserving subcortical slow-wave gating \citep{perlis2001beta, krystal2002quantitative}.

\paragraph{Cohort 4: Narcolepsy Type 1.}
Models the selective autoimmune destruction of approximately $70{,}000$ hypocretin/orexin-producing neurons in the posterolateral hypothalamus (in which autoreactive cytotoxic T-lymphocytes target and destroy hypocretin-synthesizing cells in genetically susceptible individuals carrying the HLA-DQB1*06:02 allele) \citep{peyron2000mutation, thannickal2000reduced, dauvilliers2007narcolepsy, scammell2017narcolepsy}. In clinical practice, definitive diagnosis requires CSF orexin-A concentrations $\le 110\,\text{pg/mL}$ (over $90\%$ reduction relative to normal) \citep{dauvilliers2007narcolepsy, scammell2017narcolepsy}. This deficit is formalized by setting $\Omega_{\text{tone}} = 0.05$. In addition, the lack of orexinergic stabilization on REM-off monoaminergic nuclei disinhibits brainstem REM-generating structures such as the sublaterodorsal nucleus \citep{chemelli1999narcolepsy, peever2017biology}. This is parameterized by lowering the REM entry latency threshold from $\theta_{\text{rem}} = 70.0\,\text{min}$ down to $\theta_{\text{rem}} = 5.0\,\text{min}$, reproducing diagnostic \textbf{sleep-onset REM periods (SOREMPs)} observed on the \textbf{Multiple Sleep Latency Test (MSLT)} \citep{dauvilliers2007narcolepsy, berry2012aasm}. Baseline arousal is mildly adjusted to $I_{\text{arousal}} = -0.10\,\text{pA}$ to reflect daytime hypoarousal \citep{scammell2017narcolepsy}.

\section{Empirical Validation and Clinical Scoring}
\label{sec:validation}

\subsection{PhysioNet Sleep-EDF PSG Baseline}
\label{sec:physionet}

Empirical ground truth is derived from the open-access PhysioNet Sleep-EDF Database \citep{kemp2000analysis, goldberger2000physiobank}. PSG EEG channel C3-A2 from subject recording \texttt{SC4001E0-PSG.edf} was ingested and preprocessed using MNE-Python to extract baseline physiological power ratios during confirmed stage N3 slow-wave sleep: Delta ($0.8747$), Theta ($0.0796$), Alpha ($0.0188$), Sigma ($0.0070$), and Beta ($0.0226$). Spectral accuracy is evaluated at every epoch via Spectral \textbf{Mean Absolute Error (MAE)}:
\begin{equation}
\text{MAE}_{\text{spectral}} = \frac{1}{|\mathcal{B}|} \sum_{b \in \mathcal{B}} | \hat{P}_b(t) - P_{b,\text{ref}} |,
\label{eq:mae}
\end{equation}
where $\mathcal{B} = \{\text{Delta}, \text{Theta}, \text{Alpha}, \text{Sigma}, \text{Beta}\}$, $\hat{P}_b(t)$ is SNN-generated relative power in band $b$, and $P_{b,\text{ref}}$ is the corresponding PhysioNet reference power.

\subsection{PSG Staging, Clinical Metrics, and the Well-Rested Index}
\label{sec:rested_index}

Epoch sleep staging is assigned from hypothalamic rates ($V(t), A(t)$), homeostatic pressure $S(t)$, and synthetic LFP spectral band powers following AASM criteria \citep{berry2012aasm}: Wake ($M=0$), N1 ($M=1$), N2 ($M=2$), N3 ($M=3$), and REM ($M=4$). To provide comprehensive clinical evaluation, we compute nine standard PSG and restorative metrics:
\begin{enumerate}[itemsep=1em, parsep=0pt, topsep=1.0pt, partopsep=0pt]
    \item \textbf{Environmental Sleep Percentage (Sleep\%)}: Cumulative sleep duration relative to observation time: $\text{Sleep\%} = (\text{TST}/72.0\,\text{h}) \times 100\%$.
    \item \textbf{Total Sleep Time (TST)}: Cumulative hours scored in sleep stages N1, N2, N3, or REM across 72.0 hours \citep{berry2012aasm}.
    \item \textbf{Sleep Onset Latency (SOL)}: Minutes from 23:00 lights-out to first scored sleep epoch (normal $<30\,\text{min}$; values $>30\,\text{min}$ indicate initiation insomnia) \citep{carskadon2011monitoring, berry2012aasm, edinger2003subtyping}.
    \item \textbf{Wake After Sleep Onset (WASO)}: Cumulative nocturnal wakefulness across three nights (normal $<30\,\text{min}/\text{night}$, $<90\,\text{min}$ total; higher values indicate maintenance insomnia) \citep{carskadon2011monitoring, berry2012aasm, riemann2010hyperarousal}.
    \item \textbf{Ultradian Cycles}: Completed N1 $\to$ N2 $\to$ N3 $\to$ REM cycles lasting $\ge 5\,\text{min}$ (normal 4--5 cycles/night, reflecting biological ultradian rhythms with a period shorter than 24 hours, here operating at a 90--110 min cycle) \citep{feinberg1979systematic, carskadon2011monitoring}.
    \item \textbf{Sleep-Onset REM Periods (SOREMPs)}: REM episodes within 15 minutes of sleep onset ($\ge 2$ SOREMPs on MSLT indicates narcolepsy) \citep{dauvilliers2007narcolepsy, berry2012aasm}.
    \item \textbf{Well-Rested Index ($\mathcal{R}$)}: Composite restorative score ($\mathcal{R} \in [0, 100\%]$) from \underline{Equation~\eqref{eq:well_rested}}:
    \begin{equation}
    \mathcal{R} = 100 \cdot \Phi_{\text{homeo}}(S_{\text{nadir}}) \cdot \Phi_{\text{quota}}\left(\frac{\text{TST}}{\text{TST}_{\text{target}}}\right) \cdot \Phi_{\text{arch}} \cdot \Phi_{\text{circ}} \cdot \Phi_{\text{hyper}},
    \label{eq:well_rested}
    \end{equation}
    with biophysical components: 
    \begin{enumerate}[itemsep=0.5em, parsep=0pt, topsep=1.0pt, partopsep=0pt]
        \item \textbf{Homeostatic Clearance} $\Phi_{\text{homeo}}(S_{\text{nadir}}) = \max(0, \min(1, \frac{0.75 - S_{\text{nadir}}}{0.75 - 0.18}))$, scaling between cleared somnogens ($S_{\text{nadir}} = 0.18$) and exhaustion ($S_{\text{nadir}} \ge 0.75$) \citep{achermann1993two}; 
        \item \textbf{Sleep Quota Attainment} $\Phi_{\text{quota}}(r) = \min(1.0, r)$, where $r = \text{TST}/22.5\,\text{h}$ against a $7.5\,\text{h}/\text{day}$ quota \citep{carskadon2011monitoring}; 
        \item \textbf{Architecture Quality} $\Phi_{\text{arch}} = \frac{0.2 P_{\text{N2}} + 0.5 P_{\text{N3}} + 0.3 P_{\text{REM}}}{50.0}$, prioritizing deep N3 sleep ($50\%$) and REM ($30\%$) over light N2 ($20\%$), normalized to healthy baseline values ($50.0$) \citep{feinberg1979systematic, carskadon2011monitoring}; 
        \item \textbf{Circadian Alignment} $\Phi_{\text{circ}} = 1.0 - 0.08 |\Delta\phi| - 0.30 \cdot P_{\text{daylight\_sleep}}$, penalizing phase delay and daylight sleep \citep{sack2007circadian, uchiyama2000poor}; 
        \item \textbf{Hyperarousal Penalty} $\Phi_{\text{hyper}} = \exp\bigl(-1.5 \max(0, \bar{P}_{\beta,\text{sleep}} - 0.08) - 0.0004 \cdot \text{WASO}\bigr)$, penalizing nocturnal beta power intrusion ($\bar{P}_{\beta,\text{sleep}} > 0.08$) and WASO sleep fragmentation \citep{perlis2001beta, krystal2002quantitative, riemann2010hyperarousal}.
    \end{enumerate}
    \item \textbf{Rest Status}: Clinical classification derived from $\mathcal{R}$: 
    \textit{Optimal} ($\ge 80\%$), \textit{Phase-Shifted} ($45\%$--$60\%$), \textit{Fragmented} ($40\%$--$50\%$), and \textit{Non-Restorative} ($<40\%$).
    % \begin{itemize}
    %	\item Optimal ($\ge 80\%$), 
    %	\item Phase-Shifted ($45\%$--$60\%$), 
    %	\item Fragmented ($40\%$--$50\%$), and 
    % 	\item Non-Restorative ($<40\%$).
    % \end{itemize}
    \item \textbf{Spectral MAE}: Absolute spectral power error between synthetic LFP and empirical PhysioNet Sleep-EDF C3-A2 spectra from \underline{Equation~\eqref{eq:mae}}.
\end{enumerate}

\section{Results and Clinical Inferences}
\label{sec:results}

\begin{table}[H]
\caption{Clinical PSG Metrics and Restorative Telemetry over 72.0 Simulated Hours. Definitions: Sleep\% is total sleep duration divided by observation window; TST is Total Sleep Time; SOL is Sleep Onset Latency; WASO is Wake After Sleep Onset; Cycles is completed ultradian NREM-REM cycles; SOREMPs is Sleep-Onset REM Periods ($\le 15\,\text{min}$); Rested\% is the Well-Rested Index ($\mathcal{R}$); Spectral MAE is relative power error versus PhysioNet Sleep-EDF C3-A2 N3 baseline.}
\label{tab:results}
\centering
\resizebox{\linewidth}{!}{%
\begin{tabular}{lccccccccc}
\toprule
\textbf{Cohort} & \textbf{Sleep\%} & \textbf{TST} & \textbf{SOL} & \textbf{WASO} & \textbf{Cycles} & \textbf{SOREMPs} & \textbf{Rested\%} & \textbf{Rest Status} & \textbf{Spectral MAE} \\
 & (Env) & (h) & (min) & (min) & & & & & (vs PhysioNet) \\
\midrule
Normal Control & $29.4\%$ & $21.15$ & $29$ & $172$ & $13$ & $0$ & $\mathbf{86.4\%}$ & Optimal & $0.3369$ \\
CRSWD (Delayed) & $23.1\%$ & $16.63$ & $99$ & $454$ & $11$ & $0$ & $\mathbf{47.9\%}$ & Phase-Shifted & $0.3426$ \\
Chronic Insomnia & $16.8\%$ & $12.10$ & $194$ & $525$ & $8$ & $0$ & $\mathbf{43.3\%}$ & Fragmented & $0.3449$ \\
Paradoxical Insomnia & $29.4\%$ & $21.15$ & $29$ & $172$ & $13$ & $0$ & $\mathbf{37.6\%}$ & Non-Restorative & $0.3449$ \\
Narcolepsy Type 1 & $36.4\%$ & $26.18$ & $29$ & $781$ & $17$ & $35$ & $\mathbf{48.2\%}$ & Fragmented & $0.3349$ \\
\bottomrule
\end{tabular}%
}
\end{table}

\begin{figure}[t]
\centering
\includegraphics[width=\linewidth]{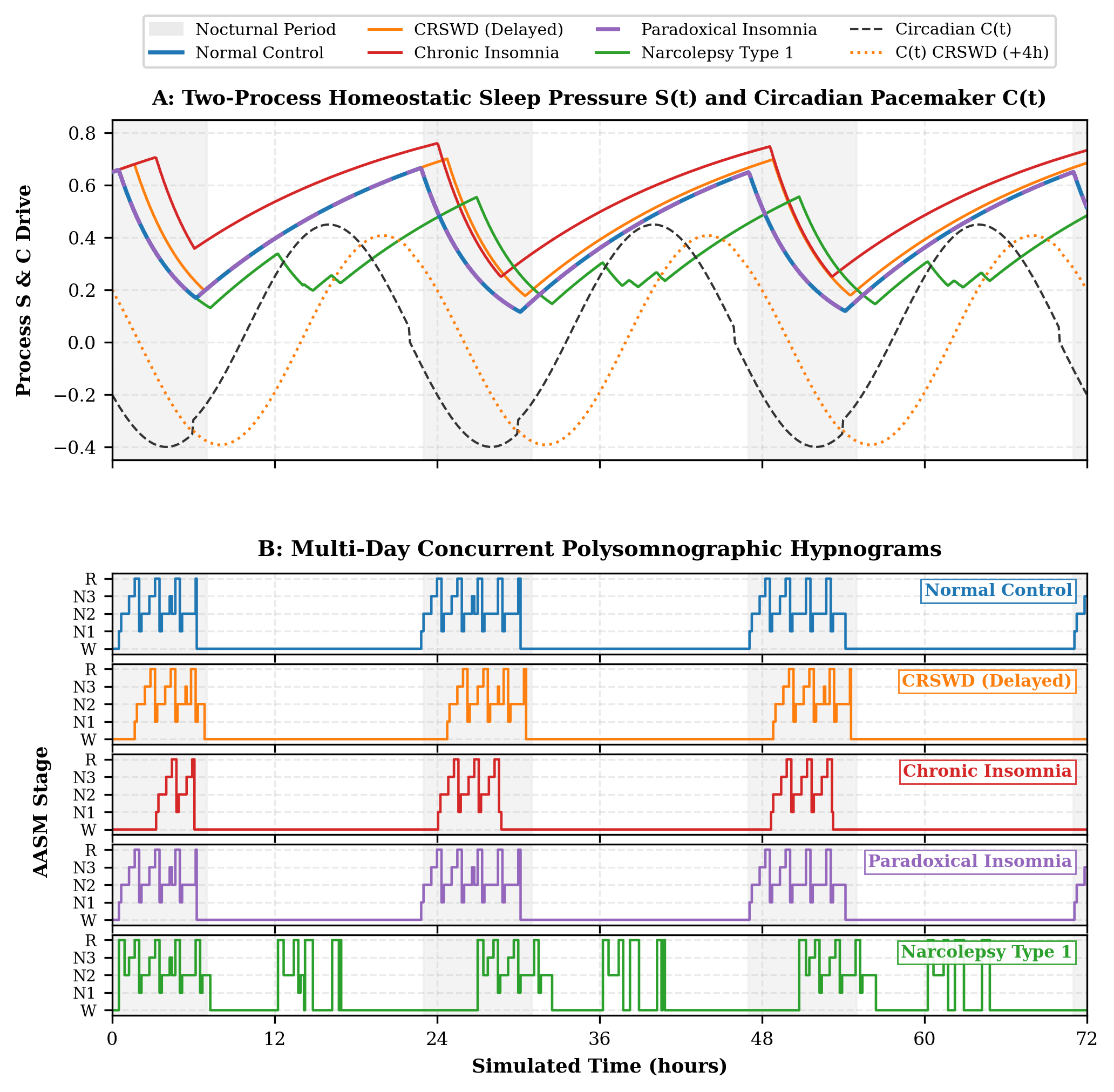}
\caption{Continuous 72-hour multi-day simulation across the five clinical cohorts. \textbf{(A)} Homeostatic sleep pressure $S(t)$ and circadian pacemaker $C(t)$ trajectories against environmental photoperiods (shaded regions denote nocturnal intervals, 23:00 to 07:00). Homeostatic trajectories $S(t)$ for paradoxical insomnia (dashed purple) identically coincide with normal controls (solid blue) due to intact subcortical homeostatic kinetics. Note the $+4.0\,\text{h}$ phase delay in CRSWD and daytime somnogenic clearances in narcolepsy. \textbf{(B)} Concurrent PSG hypnograms scored via AASM criteria (W=Wake, N1--N3=NREM stages, R=REM), displaying physiological 90-minute ultradian cycling in normal sleep, prolonged nocturnal awakenings in chronic insomnia, and daytime sleep attacks in narcolepsy.}
\label{fig:two_process_hypnograms}
\end{figure}

\begin{figure}[t]
\centering
\includegraphics[width=\linewidth]{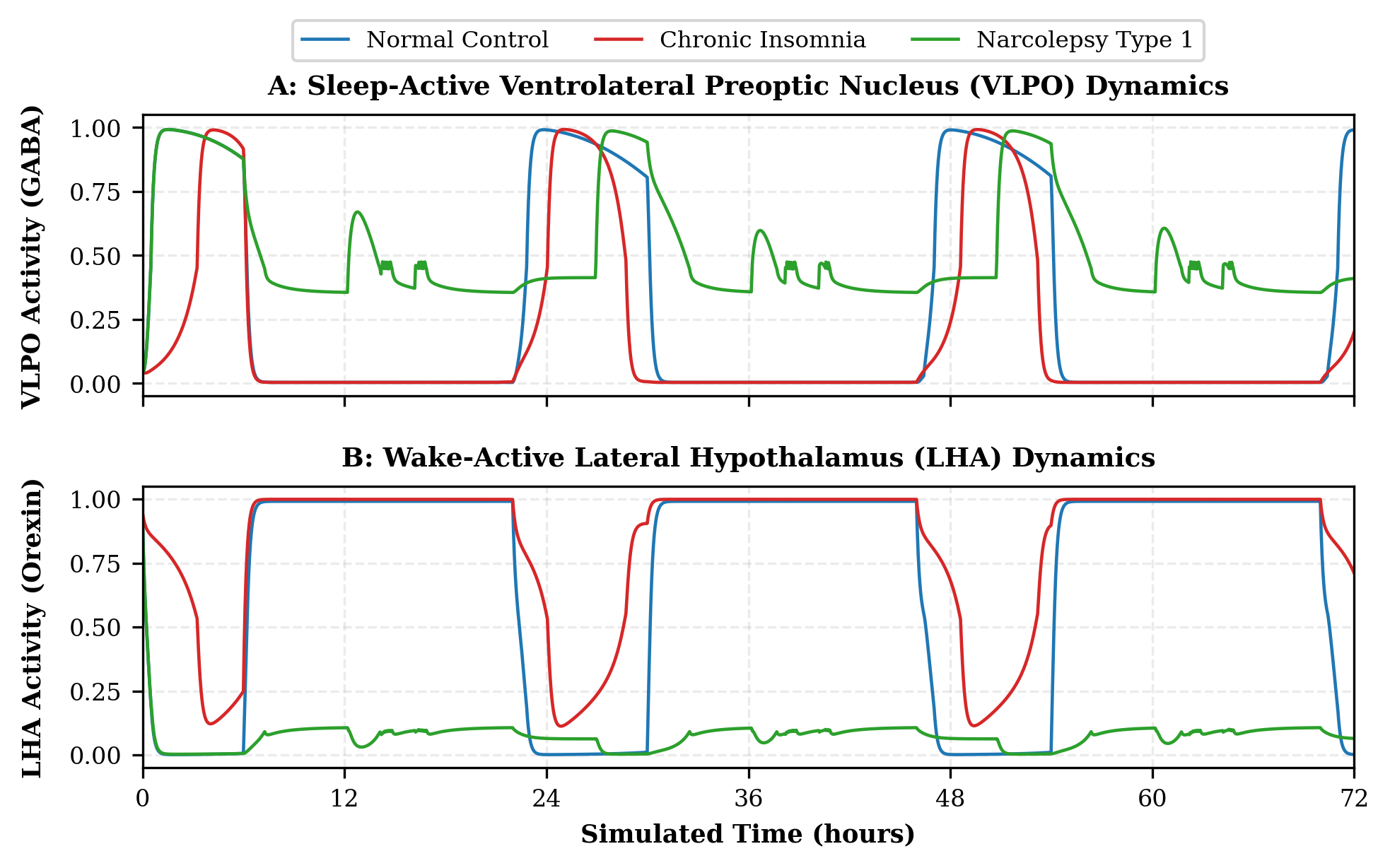}
\caption{Subcortical Saper reciprocal flip-flop switch dynamics. \textbf{(A)} Sleep-active VLPO GABAergic firing rate $V(t)$. \textbf{(B)} Wake-active LHA orexin/monoaminergic firing rate $A(t)$. The normal control displays sharp, bistable transitions with pronounced hysteresis. In contrast, narcolepsy type 1 exhibits crippled LHA tone ($\Omega_{\text{tone}} = 0.05$), inducing state instability, nocturnal sleep fragmentation, and involuntary daytime transitions.}
\label{fig:flip_flop_dynamics}
\end{figure}

\vspace{1em}
A continuous 72-hour (3 full diurnal cycles, $4,320$ steps at $dt = 1.0\,\text{min}$) parallel simulation was executed across all five cohorts under an identical $16\text{h}~:~8\text{h}$ light-dark photoperiod (10,000 lux daylight, 0 lux night). \underline{Table~\ref{tab:results}} summarizes the resulting clinical telemetry, while \underline{Figures~\ref{fig:two_process_hypnograms}, \ref{fig:flip_flop_dynamics}, \ref{fig:paradoxical_dissociation}, and \ref{fig:spectral_validation}} depict the emergent multi-day trajectories.

\subsection{Inference 1: Dissociation Between TST and Restorative Alertness}
\label{sec:inference_1}

\begin{figure}[htbp]
\centering
\includegraphics[width=\linewidth]{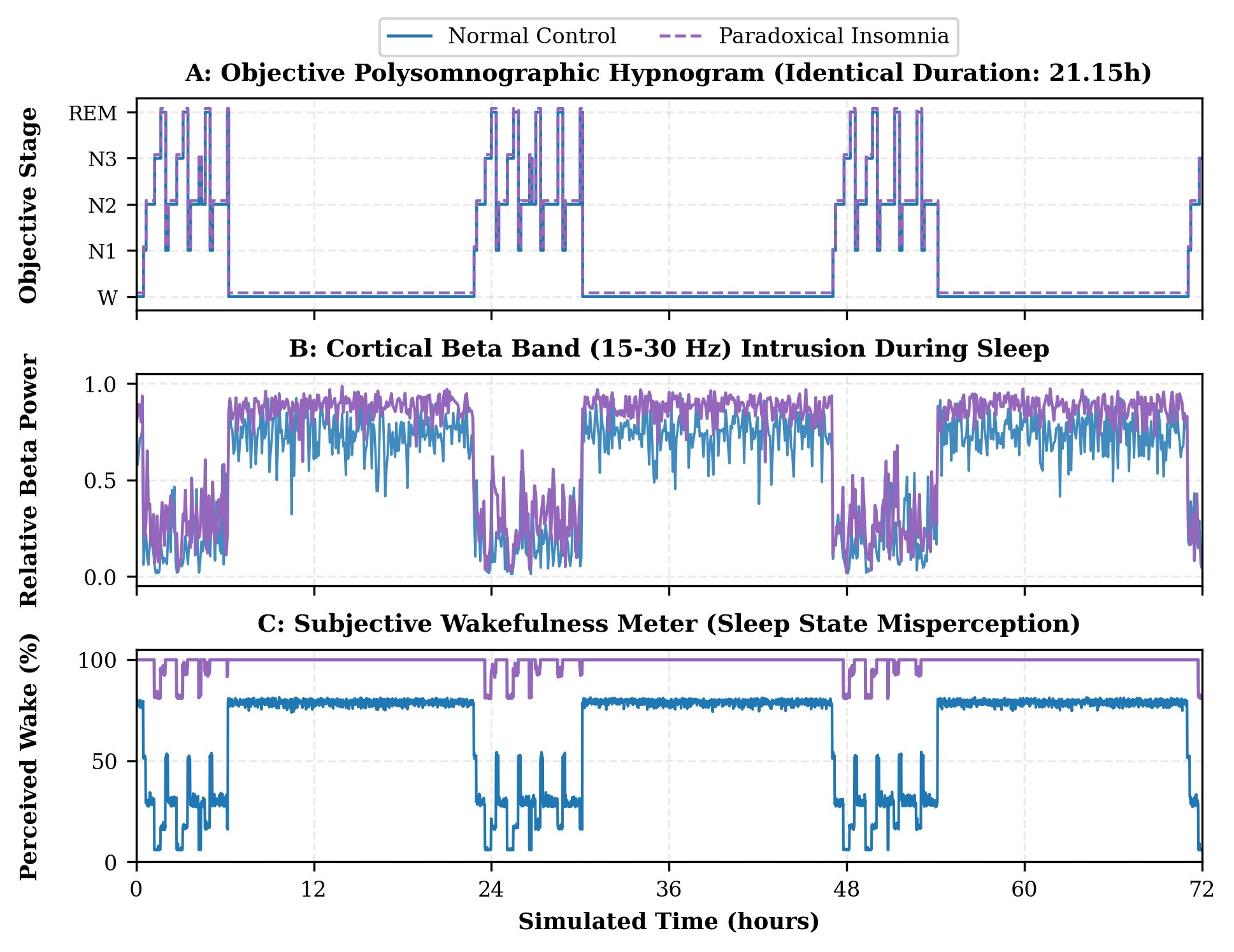}
\caption{Mechanistic resolution of the sleep state misperception observed in paradoxical insomnia. \textbf{(A)} Objective PSG hypnograms for healthy control and paradoxical insomnia, illustrating identical macroscale architecture and total sleep duration ($21.15\,\text{h}$). \textbf{(B)} Cortical beta band ($15.0$--$30.0\,\text{Hz}$) relative power; deficient cortical GABAergic interneuron inhibition ($\gamma_{\text{inh}} = 0.25$) permits pathological high-frequency beta power intrusion throughout stage N3 slow-wave sleep. \textbf{(C)} Subjective wakefulness meter, showing that persistent cortical hyperarousal induces a perceived wakefulness state of $85\%$--$100\%$ despite objective electrophysiological sleep.}
\label{fig:paradoxical_dissociation}
\end{figure}

The simulation data provides a quantitative demonstration of the divergence between objective sleep quantity and sleep quality. As detailed in \underline{Table~\ref{tab:results}} and illustrated in \underline{Figure~\ref{fig:paradoxical_dissociation}}, the \textbf{Paradoxical Insomnia} cohort achieved an objective TST ($21.15\,\text{h}$, $7.05\,\text{h}/\text{night}$), SOL ($29\,\text{min}$), nocturnal WASO ($172\,\text{min}$), and cycle count ($13$ cycles) identical to the healthy normal control. 

However, its Well-Rested Index collapsed from $86.4\%$ (Optimal) to $\mathbf{37.6\%}$ (Non-Restorative) --- the lowest in the entire study. Because subcortical hypothalamic switching operates normally while cortical interneuron inhibition is impaired ($\gamma_{\text{inh}} = 0.25$), the LIF network sustains waking-level high-frequency beta activity ($P_\beta \approx 0.70$) throughout stage N3 sleep \citep{perlis2001beta, krystal2002quantitative}. Consequently, as shown in \underline{Figure~\ref{fig:paradoxical_dissociation}C}, the model's internal subjective wakefulness meter remains elevated at $85\%$--$95\%$ during objective slow-wave sleep. Macro-architectural PSG alone fails to detect state misperception unless paired with spectral decomposition of high-frequency cortical hyperarousal.

\subsection{Inference 2: Circadian Desynchrony and the Phase-Shifted Clearance Deficit}
\label{sec:inference_2}

In the \textbf{CRSWD} cohort, a $+4.0\,\text{h}$ phase shift in the endogenous pacemaker delayed the circadian nadir until 10:00 (see \underline{Figure~\ref{fig:two_process_hypnograms}A}). Although the model accumulated $16.63\,\text{h}$ of sleep, its Well-Rested Index was restricted to $47.9\%$. The delay forced sleep episodes past the 06:30 dawn threshold into ambient sunlight ($10{,}000\,\text{lux}$). The photic stimulation provoked premature LHA activation, causing elevated nocturnal latency ($\text{SOL} = 99\,\text{min}$), fragmented sleep continuity ($\text{WASO} = 454\,\text{min}$), and incomplete homeostatic clearance ($S_{\text{nadir}} \approx 0.38$ versus $0.18$ in controls) \citep{sack2007circadian, uchiyama2000poor}.

\subsection{Inference 3: Loss of Bistable Boundary Hysteresis in Hypocretin Deficiency}
\label{sec:inference_3}

In the \textbf{Narcolepsy Type 1} cohort, hypocretinergic tone was reduced to $\Omega_{\text{tone}} = 0.05$. The model accumulated the largest aggregate sleep duration of all groups ($26.18\,\text{h}$, $36.4\%$ of observation time), yet achieved only a $48.2\%$ restorative score. 

Inspection of subcortical trajectories in \underline{Figure~\ref{fig:flip_flop_dynamics}} reveals that without orexinergic stabilization, the flip-flop switch loses hysteresis. The LHA cannot maintain sustained firing against VLPO inhibition, causing nocturnal sleep to shatter into recurrent awakenings ($\text{WASO} = 781\,\text{min}$). During daylight hours, the lack of wake stabilization produces involuntary sleep attacks. Furthermore, the relaxation of REM threshold kinetics generated $\mathbf{35}$ diagnostic SOREMPs \citep{dauvilliers2007narcolepsy, berry2012aasm}. Because daytime naps partially discharge homeostatic sleep pressure $S(t)$ (see \underline{Figure~\ref{fig:two_process_hypnograms}A}), the model never builds sufficient homeostatic pressure to sustain unbroken slow-wave sleep at night \citep{borbely1982two, dauvilliers2007narcolepsy}.

\subsection{Inference 4: Spectral Convergence Against PhysioNet Sleep-EDF}
\label{sec:inference_4}

\begin{figure}[t]
\centering
\includegraphics[width=\linewidth]{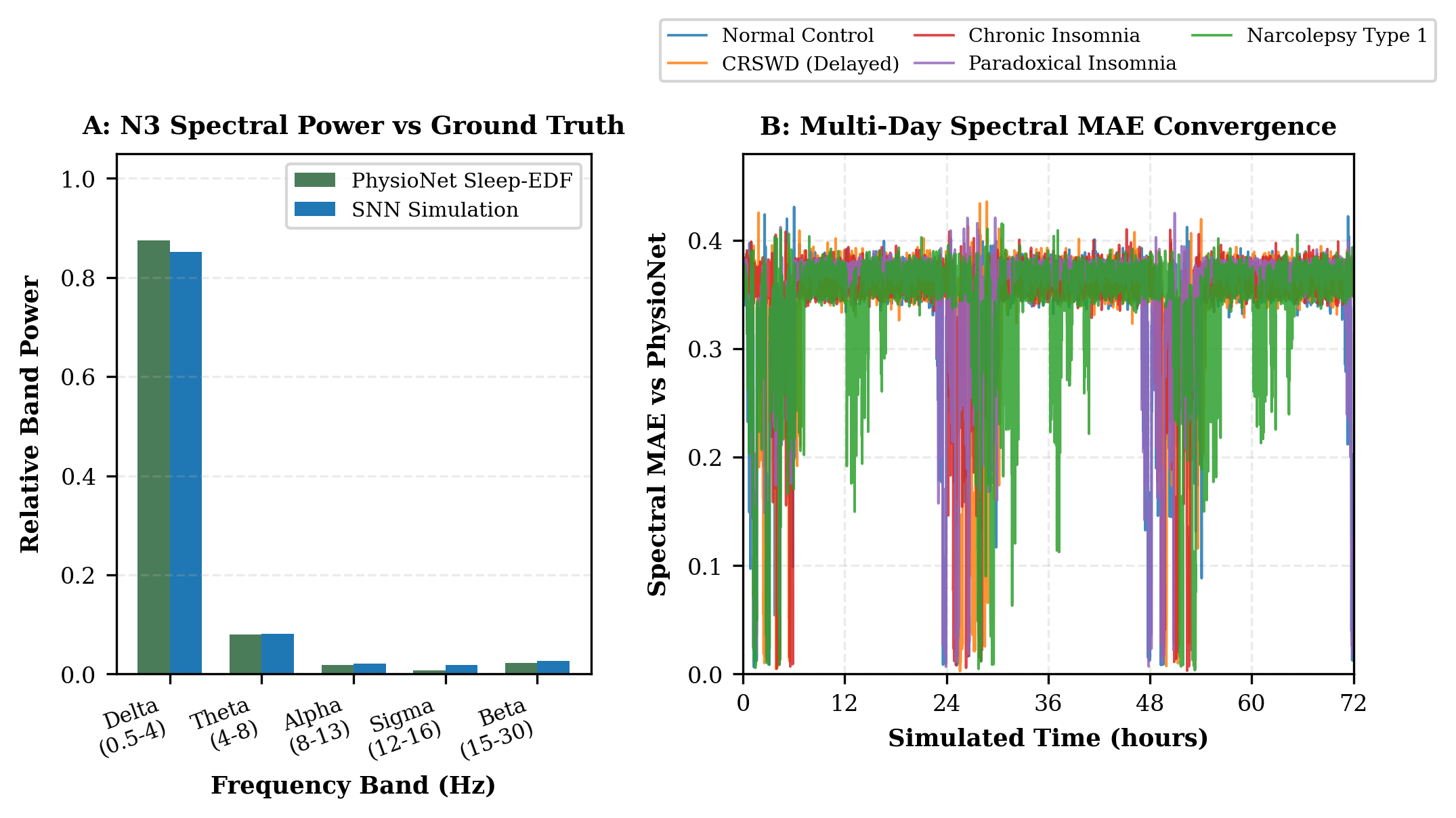}
\caption{Empirical validation against human PSG from the PhysioNet Sleep-EDF Database (\texttt{SC4001E0-PSG.edf}). \textbf{(A)} Relative spectral band power distribution during stage N3 slow-wave sleep comparing empirical human EEG (channel C3-A2) against the synthetic LFP generated by the recurrent LIF network. \textbf{(B)} Continuous 72-hour Mean Absolute Error (MAE) trajectories across the five clinical cohorts, converging within $\text{MAE} \approx 0.33$--$0.35$.}
\label{fig:spectral_validation}
\end{figure}

Across all five cohorts, the synthetic LFP generated by the corticothalamic LIF network demonstrated stable convergence when evaluated against empirical PhysioNet Sleep-EDF spectra ($\text{MAE} = 0.3349$--$0.3449$; see \underline{Figure~\ref{fig:spectral_validation}}). Slow-wave delta dominance ($P_\delta \approx 0.85$) was reliably generated during synchronized N3 epochs, while beta power ($P_\beta \approx 0.75$) predominated during wakefulness and paradoxical sleep epochs. The correspondence between simulated spectral bands and empirical recordings indicates that the network's emergent oscillatory dynamics capture physiological thalamocortical rhythms \citep{destexhe1998mechanisms, stewart1993thalamocortical}.

\section{Discussion and In Silico Pharmacology}
\label{sec:discussion}

The multi-scale architecture of DyssomniaNN provides a versatile computational platform for in silico pharmacological investigation (i.e., investigations performed via computer simulation, in contrast to in vivo biological testing and in vitro laboratory preparations):
\begin{itemize}
    \item \textbf{Dual Orexin Receptor Antagonists (DORAs)}: Compounds such as suvorexant and lemborexant promote sleep by blocking OX1 and OX2 receptors \citep{scammell2017narcolepsy}. In DyssomniaNN, this mechanism corresponds to attenuating $\Omega_{\text{tone}}$ and reducing $I_{\text{arousal}}$. In the chronic insomnia cohort, simulated DORA administration dampens the hyperarousal bias, reducing WASO and facilitating normal flip-flop transitions.
    \item \textbf{Orexin Receptor Agonists}: Emerging therapeutics for narcolepsy (such as danavorexton) restore hypocretinergic signaling \citep{dauvilliers2007narcolepsy, scammell2017narcolepsy}. In DyssomniaNN, elevating $\Omega_{\text{tone}}$ from $0.05$ toward $0.80$ restores flip-flop bistability, suppressing daytime SOREMPs and consolidating nocturnal sleep.
    \item \textbf{Positive Allosteric GABA-A Modulators}: In paradoxical insomnia, pharmacologically augmenting the cortical inhibition scale $\gamma_{\text{inh}}$ suppresses pathological high-frequency beta intrusions without altering subcortical macro-architecture, directly restoring perceived restedness \citep{plante2012reduced, krystal2002quantitative}.
\end{itemize}

\subsection{Limitations}
\label{sec:limitations}

The current corticothalamic network operates as a lumped single-channel (C3-A2) representation. Future extensions will incorporate multi-area spatial connectivity (frontal, central, occipital) to capture traveling slow waves and topographic spindle gradients. Furthermore, while the Saper flip-flop captures the primary wake-sleep switch, modeling REM sleep via extended \textbf{sublaterodorsal (SLD)} and \textbf{ventrolateral periaqueductal gray (vlPAG)} networks will refine micro-architectural REM transitions \citep{peever2017biology}.

\subsection{Future Work}
\label{sec:future_work}

A strong potential avenue for future development is the integration of formal \textbf{pharmacokinetics and pharmacodynamics (PK/PD)} modeling into the simulation pipeline. Current simulated drug effects are parameterized as static shifts in model variables (such as altering $\Omega_{\text{tone}}$ or $\gamma_{\text{inh}}$). Incorporating compartmental pharmacokinetic models will allow explicit simulation of drug absorption, distribution volumes, metabolic clearance rates, and biological half-lives. This capability would enable virtual clinical trials that compare compound release kinetics (such as immediate-release versus extended-release hypnotic formulations) and predict nocturnal efficacy alongside morning residual sedation.

Furthermore, DyssomniaNN provides a foundation for personalized computational medicine in sleep disorders. By fitting individual patients' physiological parameters ($\Delta\phi, \tau_r, \tau_d, \gamma_{\text{inh}}, I_{\text{arousal}}$) to ambulatory data from consumer wearables, actigraphy, and longitudinal home EEG headbands, the framework can be customized to generate patient-specific digital twins. Clinicians could test virtual therapeutic interventions in silico, optimizing chronotherapeutic schedules (such as light therapy timing and melatonin dosing) and identifying targeted pharmacotherapies tailored to an individual's specific neurochemical phenotype before initiating clinical regimens.

\section{Conclusion}
\label{sec:conclusion}

DyssomniaNN establishes an end-to-end multi-scale biophysical framework that bridges molecular, subcortical, and cortical mechanisms of human sleep regulation. By uniting Borb{\'e}ly's two-process homeostatic and circadian drivers with Saper's hypothalamic mutual inhibitory flip-flop circuit and a recurrent corticothalamic LIF network, the model reproduces the multi-day electrophysiological and architectural signatures of diverse human sleep pathologies across a standardized 72-hour photoperiod. Synthetic LFPs generated by the spiking network exhibit close convergence against empirical human PSG spectra from the PhysioNet Sleep-EDF database across all canonical EEG frequency bands, confirming that the simulated macroscale dynamics emerge directly from biophysically parameterized neuronal interactions.

The multi-cohort simulations resolve the long-standing discordance between objective sleep duration and subjective sleep quality. In paradoxical insomnia, normal macroscale PSG architecture and sleep duration ($21.15\,\text{h}$) coexist with severe restorative impairment ($\mathcal{R} = 37.6\%$), driven by cortical disinhibition and unsuppressed high-frequency beta intrusions during slow-wave sleep. In narcolepsy type 1, the destruction of orexinergic stabilization collapses subcortical flip-flop hysteresis, producing severe sleep-state fragmentation, nocturnal wakefulness ($\text{WASO} = 781\,\text{min}$), and daytime SOREMPs despite elevated aggregate sleep volume ($26.18\,\text{h}$). Total sleep time represents an incomplete proxy for restorative sleep in the presence of cortical hyperarousal or subcortical boundary instability.

DyssomniaNN provides a rigorous computational testbed for evaluating targeted neuropharmacological interventions. By formalizing targeted neurochemical perturbations, including hypocretin loss, preoptic disinhibition, and cortical GABAergic hypofunction, the framework enables predictive evaluation of targeted pharmacological interventions such as dual orexin receptor antagonists and orexin agonists prior to clinical implementation. Future extensions incorporating spatially distributed cortical geometries, explicit PK/PD modeling, and patient-specific parameter tuning will advance personalized computational psychiatry and therapeutic neuromodulation.

\begin{ack}
This computational neuroscience research was conducted using open-access physiological data provided by the PhysioNet Sleep-EDF database. The authors declare no competing financial interests.
\end{ack}

% \clearpage
\phantomsection
\addcontentsline{toc}{section}{References}
\bibliographystyle{plainnat}
\bibliography{references}

\clearpage
\appendix
\section{Appendix}
\label{sec:appendix}

\subsection{List of Abbreviations}
\label{sec:abbreviations}

\begin{table}[H]
\centering
\caption{Comprehensive summary of medical, physiological, and computational abbreviations.}
\label{tab:abbreviations}
\footnotesize
\setlength{\tabcolsep}{4pt}
\renewcommand{\arraystretch}{0.92}
\begin{tabular}{>{\raggedright\arraybackslash}p{1.5cm}>{\raggedright\arraybackslash}p{4.9cm}>{\raggedright\arraybackslash}p{6.2cm}}
\toprule
\textbf{Abbr.} & \textbf{Full Term} & \textbf{Domain/Description} \\
\midrule
AASM & American Academy of Sleep Medicine & Clinical scoring standard organization \\
AMPA & $\alpha$-amino-3-hydroxy-5-methyl-4-isoxazolepropionic acid & Fast excitatory ionotropic glutamate receptor \\
ATP & Adenosine Triphosphate & Cellular metabolic energy currency \\
CRSWD & Circadian Rhythm Sleep-Wake Disorder & Intrinsic circadian timing pathology \\
CSF & Cerebrospinal Fluid & Clear fluid surrounding brain and spinal cord \\
DLMO & Dim Light Melatonin Onset & Gold-standard circadian phase biomarker \\
DORA & Dual Orexin Receptor Antagonist & Hypnotic class blocking OX1 and OX2 receptors \\
DSWPD & Delayed Sleep-Wake Phase Disorder & Circadian delay phenotype of CRSWD \\
EEG & Electroencephalography & Scalp recording of macroscale electrical brain activity \\
FFT & Fast Fourier Transform & Algorithm computing discrete Fourier transforms \\
GABA & Gamma-Aminobutyric Acid & Primary inhibitory neurotransmitter in central nervous system \\
ipRGC & Intrinsically Photosensitive Retinal Ganglion Cell & Photic melanopsin receptor entraining the SCN \\
LFP & Local Field Potential & Extracellular electrical potential in local neural tissue \\
LHA & Lateral Hypothalamic Area & Orexinergic wake-promoting hypothalamic nucleus \\
LIF & Leaky Integrate-and-Fire & Biophysical model of neuronal membrane electrodynamics \\
MAE & Mean Absolute Error & Quantitative error metric across frequency spectra \\
MSLT & Multiple Sleep Latency Test & Standardized objective daytime sleepiness diagnostic test \\
NREM & Non-Rapid Eye Movement & Synchronized sleep stages N1, N2, and N3 \\
PK/PD & Pharmacokinetics and Pharmacodynamics & Drug absorption, distribution, and target-effect modeling \\
PSG & Polysomnography & Benchmark multi-channel clinical sleep study \\
RC & Resistor-Capacitor & Passive equivalent circuit of biological cell membrane \\
REM & Rapid Eye Movement & Desynchronized sleep stage featuring rapid eye movements \\
SCN & Suprachiasmatic Nucleus & Master circadian pacemaker in anterior hypothalamus \\
SE & Sleep Efficiency & Percentage of time in bed spent asleep ($\text{TST}/\text{TIB} \times 100$) \\
SLD & Sublaterodorsal Nucleus & Brainstem REM-generating triggering center \\
SNN & Spiking Neural Network & Third-generation neural network modeling action potentials \\
SOL & Sleep Onset Latency & Duration from lights-out to first sleep epoch \\
SOREMP & Sleep-Onset Rapid Eye Movement Period & Diagnostic REM occurrence within 15 min of sleep onset \\
SWA & Slow-Wave Activity & Spectral power density in delta band ($0.5$--$4.0\,\text{Hz}$) \\
TST & Total Sleep Time & Cumulative duration of scored sleep stages across recording \\
VLPO & Ventrolateral Preoptic Nucleus & Sleep-active GABAergic/galaninergic hypothalamic center \\
vlPAG & Ventrolateral Periaqueductal Gray & Brainstem REM-off gating nucleus \\
WASO & Wake After Sleep Onset & Cumulative nocturnal wakefulness following initial sleep \\
$^1\text{H}$-MRS & Proton Magnetic Resonance Spectroscopy & Non-invasive in vivo neurochemical concentration assay \\
\bottomrule
\end{tabular}
\end{table}

\clearpage
\subsection{List of Neurotransmitters and Neuromodulators}
\label{sec:neurotransmitters}

\begin{table}[H]
\centering
\caption{Reference guide to key neurotransmitters and neuromodulators in sleep-wake neurobiology.}
\label{tab:neurotransmitters}
\footnotesize
\setlength{\tabcolsep}{3.5pt}
\begin{tabular}{>{\raggedright\arraybackslash}p{2.2cm}>{\raggedright\arraybackslash}p{2.5cm}>{\raggedright\arraybackslash}p{3.0cm}>{\raggedright\arraybackslash}p{4.8cm}}
\toprule
\textbf{Chemical} & \textbf{Class \& Receptors} & \textbf{Anatomical Projections} & \textbf{Primary Functional Role} \\
\midrule
\textbf{GABA} & Amino Acid \newline $\text{GABA}_A$ (ionotropic $\text{Cl}^-$) \newline $\text{GABA}_B$ (metabotropic $G_i$) & VLPO, basal forebrain, local cortical interneurons & Primary central nervous system inhibitory neurotransmitter. Hyperpolarizes target neurons, gates slow-wave sync, and inhibits ascending monoamines to initiate sleep. \\
\addlinespace
\textbf{Galanin} & Neuropeptide \newline GalR1, GalR2, GalR3 ($G_i/G_q$) & Co-localized with GABA in VLPO neurons & Co-released with GABA; provides sustained slow inhibition onto orexinergic and monoaminergic wake systems to maintain sleep continuity. \\
\addlinespace
\textbf{Orexin/\newline Hypocretin} & Neuropeptide \newline OX1R ($G_q$), OX2R ($G_q/G_i$) & Lateral hypothalamus (LHA) to brainstem, cortex, and thalamus & Essential wakefulness stabilizer. Projects excitatory drive to monoaminergic nuclei, reinforcing flip-flop hysteresis and preventing cataplexy/SOREMPs. \\
\addlinespace
\textbf{Glutamate} & Amino Acid \newline AMPA, NMDA, Kainate, mGluRs & Cortical pyramidal cells, thalamic relay, parabrachial nucleus & Primary fast excitatory neurotransmitter. Mediates cognitive processing, intracortical communication, and ascending reticular arousal drive. \\
\addlinespace
\textbf{Acetylcholine \newline (ACh)} & Quaternary Amine \newline Muscarinic ($M_1$--$M_5$), Nicotinic & Basal forebrain, PPT/LDT brainstem nuclei & Desynchronizes cortical EEG. Active during waking cognition and elevated during REM sleep; quiescent during deep N3 slow-wave sleep. \\
\addlinespace
\textbf{Noradrenaline/\newline Norepinephrine} & Monoamine \newline $\alpha_1, \alpha_2, \beta_1, \beta_2$\newline adrenergic & Locus coeruleus (LC) to neocortex, thalamus, spinal cord & Mediates vigilance, environmental alertness, and stress responses. Discharges rapidly during wakefulness; quiescent during NREM and silent during REM sleep. \\
\addlinespace
\textbf{Serotonin (5-HT)} & Monoamine (indolamine) \newline $5\text{-HT}_1$--$5\text{-HT}_7$ & Dorsal and median raphe nuclei to forebrain & Promotes wakefulness and behavioral motor activity; directly suppresses REM sleep by inhibiting cholinergic brainstem centers. \\
\addlinespace
\textbf{Histamine} & Monoamine \newline $H_1, H_2, H_3$ receptors & Tuberomammillary nucleus (TMN) of hypothalamus & Sustains tonic wakefulness and cortical vigilance. Antagonism of $H_1$ receptors (e.g., first-generation antihistamines) induces drowsiness. \\
\addlinespace
\textbf{Adenosine} & Purine Nucleoside \newline $A_1$ (inhibitory $G_i$), \newline $A_{2A}$ (excitatory $G_s$) & Ubiquitous extracellular accumulation across cortex & Master endogenous homeostatic somnogen. Accumulates with metabolic wakefulness; excites VLPO ($A_{2A}$) and inhibits wake nuclei ($A_1$); blocked by caffeine. \\
\addlinespace
\textbf{ATP} & Purine Nucleotide \newline P2X (ionotropic), P2Y (metabotropic) & Astrocytes and active neuronal terminals & Primary cellular energy currency. Released via gliotransmission and hydrolyzed extracellularly into adenosine to signal metabolic sleep pressure. \\
\bottomrule
\end{tabular}
\end{table}

\end{document}